# Multiple Source Adaptation and the Rényi Divergence

**Yishay Mansour**
Google Research and
Tel Aviv Univ.
mansour@tau.ac.il

**Mehryar Mohri**
Courant Institute and
Google Research
mohri@cims.nyu.edu

**Afshin Rostamizadeh**
Courant Institute
New York University
rostami@cs.nyu.edu

## Abstract

This paper presents a novel theoretical study of the general problem of multiple source adaptation using the notion of Rényi divergence. Our results build on our previous work [12], but significantly broaden the scope of that work in several directions. We extend previous multiple source loss guarantees based on distribution weighted combinations to arbitrary target distributions $P$, not necessarily mixtures of the source distributions, analyze both known and unknown target distribution cases, and prove a lower bound. We further extend our bounds to deal with the case where the learner receives an approximate distribution for each source instead of the exact one, and show that similar loss guarantees can be achieved depending on the divergence between the approximate and true distributions. We also analyze the case where the labeling functions of the source domains are somewhat different. Finally, we report the results of experiments with both an artificial data set and a sentiment analysis task, showing the performance benefits of the distribution weighted combinations and the quality of our bounds based on the Rényi divergence.

## 1 Introduction

The standard analysis of *generalization* in theoretical and applied machine learning relies on the assumption that training and test points are drawn according to the same distribution. This assumption forms the basis of common learning frameworks such as the PAC model [17]. But, a number of learning tasks emerging in practice present an even more challenging generalization where the distribution of training points somewhat differs from that of the test points.

A general version of this problem is known as the *domain adaptation* problem where very few or no labeled points are available from the *target domain*, but where the learner receives a labeled training sample from a *source domain* somewhat close to the target domain and where he typically can further access a large set of unlabeled points from a *target domain*. This problem arises in a variety of natural language processing tasks such parsing, statistical language modeling, text classification [15, 7, 16, 9, 10, 4, 6], or speech processing [11, 8, 14] and computer vision [13] tasks, as well as in many other applications. Several recent studies deal with some theoretical aspects of this adaptation problem [2, 3].

A more complex variant of this problem arises in sentiment analysis and other text classification tasks where the learner receives information from *several* domain sources that he can combine to make predictions about a target domain. As an example, often appraisal information about a relatively small number of domains such as *movies*, *books*, *restaurants*, or *music* may be available, but little or none is accessible for more difficult domains such as *travel*. This is known as the *multiple source adaptation problem*. Instances of this problem can be found in a variety of other natural language and image processing tasks. A problem with multiple sources but distinct from domain adaptation has also been considered by [5] where the sources have the same input distribution but can have different labels, modulo some disparity constraints.

We recently introduced and analyzed the problem of adaptation with multiple sources [12]. The problem is formalized as follows. For each source domain $i \in [1, k]$, the learner receives the distribution of the input points $Q_i$, as well as a hypothesis $h_i$ with loss at most $\epsilon$ on that source. The task consists of combining the $k$ hypotheses $h_i, i \in [1, k]$, to derive a hypothesis $h$ with a loss as small as possible with respect to the target distribution $P$. Different scenarios can be considered according to whether the distribution $P$ is known or unknown to the learner.

We showed that solutions based on a simple convex combination of the $k$ source hypotheses $h_i$ can perform very poorly and pointed out cases where *any* such convex com-bination would incur a classification error of half, even when each source hypothesis $h_i$ makes no error on its domain $Q_i$ [12]. We proposed instead *distribution weighted combinations* of the source hypotheses, which are combinations of source hypotheses weighted by the source distributions. We showed that, remarkably, for a fixed target function, there exists a distribution weighted combination of the source hypotheses whose loss is at most $\epsilon$ with respect to *any* mixture $P$ of the $k$ source distributions $Q_i$.

This paper presents a novel theoretical study of the general problem of multiple source adaptation using the notion of *Rényi divergence* [1]. Our results build on our previous work [12], but significantly broaden the scope of that work in several ways. We extend previous multiple source loss guarantees to arbitrary target distributions $P$ not necessarily mixtures of the source distributions: we show that for a fixed target function, there exists a distribution weighted combination of the source hypotheses whose loss can be bounded with respect to the maximum loss of the source hypotheses and the Rényi divergence between the target distribution and the class of mixtures distributions.

We further extend our bounds to deal with the case where the learner receives an approximate distribution $\widehat{Q}_i$ for each source $i$ instead of the true distribution $Q_i$, and show that similar loss guarantees can be achieved depending on the divergence between the approximate and true distributions. We also analyze the case where the labeling functions $f_i$ of the source domains are somewhat different. We show that our results can be extended to tackle this situation as well, assuming that the functions $f_i$ are "close" to the target function on the target distribution, but not necessarily on the source distributions.

Much of our results are based on a family of information theoretical divergences introduced by Alfred Rényi [1], which share some of the properties of the standard relative entropy or Kullback-Leibler divergence and include it as a special case, but form an extension based on the theory of generalized means. The Rényi divergences come up naturally in our analysis to measure the distance between distributions and seem to be closely related to the adaptation generalization bounds.

The next section introduces these divergences as well as other preliminary notation and definitions. Section 3 gives general learning bounds for multiple source adaptation. This includes the analysis of both known and unknown target distribution cases, the proof of lower bounds, and the study of some natural combining rules. Section 4 presents a generalization of several of these results to the case of approximate source distributions $\widehat{Q}_i$. Section 5 presents an extension to multiple labeling functions $f_i$. Section 6 reports the results of experiments with both an artificial data set and a sentiment analysis task showing the performance benefits of the distribution weighted combinations and the quality of our bounds based on the Rényi divergence.

## 2 Preliminaries

### 2.1 Multiple Source Adaptation Problem

Let $\mathcal{X}$ be the input space, $f\colon \mathcal{X} \to \mathbb{R}$ the target function, and $L\colon \mathbb{R}\times\mathbb{R} \to \mathbb{R}$ a loss function. The loss of a hypothesis $h$ with respect to a distribution $P$ is denoted by $\mathcal{L}_P(h, f)$ and defined as $\mathcal{L}_P(h, f) = \mathrm{E}_{x\sim P}[L(h(x), f(x))] = \sum_{x\in\mathcal{X}} L(h(x), f(x)) P(x)$. We denote by $\Delta$ the simplex of $\mathbb{R}^k$: $\Delta = \{\lambda \in \mathbb{R}^k \colon \lambda_i \geq 0 \wedge \sum_{i=1}^k \lambda_i = 1\}$.

We consider an adaptation set-up with $k$ source domains and a single target domain as in [12]. The input to the problem is a target distribution $P$, a set of $k$ source distributions $Q_1, \ldots, Q_k$ and $k$ corresponding hypotheses $h_1, \ldots, h_k$ such that for all $i \in [1, k]$, $\mathcal{L}_{Q_i}(h_i, f) \leq \epsilon$, for a fixed $\epsilon \geq 0$. The adaptation problem consists of combing the hypotheses $h_i$s to derive a hypothesis with small loss on the target distribution $P$.

A *combining rule* for the hypotheses takes as an input the $h_i$s and outputs a single hypothesis $h\colon \mathcal{X} \to \mathbb{R}$. A particular combining rule introduced in [12] that we shall also use here is the *distribution weighted combining rule* which is based on a parameter $z \in \Delta$ and defines the hypothesis by $h_z(x) = \sum_{i=1}^k \frac{z_i Q_i(x)}{\sum_{j=1}^k z_j Q_j(x)} h_i(x)$ when $\sum_{j=1}^k z_j Q_j(x) > 0$ and $h_z(x) = 0$ otherwise, for all $x \in \mathcal{X}$. We denote by $\mathcal{H}$ the set of all distribution weighted combination hypotheses.

We assume that the following properties hold for the loss function $L$: (i) $L$ is non-negative: $L(x,y) \geq 0$ for all $x, y \in \mathbb{R}$; (ii) $L$ is convex with respect to the first argument: $L(\sum_{i=1}^k \lambda_i x_i, y) \leq \sum_{i=1}^k \lambda_i L(x_i, y)$ for all $x_1, \ldots, x_k, y \in \mathbb{R}$ and $\lambda \in \Delta$; (iii) $L$ is bounded: there exists $M \geq 0$ such that $L(x,y) \leq M$ for all $x, y \in \mathbb{R}$. An example of loss function verifying these assumptions is the *absolute loss* defined by $L(x,y) = |x-y|$ or the *0-1 loss*, $L_{01}$, defined for Boolean functions by $L(0,1) = L(1,0) = 1$ and $L(0,0) = L(1,1) = 0$.

### 2.2 Rényi Entropy and divergence

The *Rényi entropy* $H_\alpha$ of a distribution $P$ is parameterized by a real number $\alpha$, $\alpha > 0$ and $\alpha \neq 1$, and defined as

$$H_\alpha(P) = \frac{1}{1-\alpha} \log \sum_{x \in X} P^\alpha(x). \qquad (1)$$

For $\alpha \in \{0, 1, +\infty\}$, $H_\alpha$ is defined as the limit of $H_\lambda$ for $\lambda \to \alpha$. Let us review some specific values of $\alpha$ and the corresponding interpretation of the Rényi entropy. For $\alpha = 0$, the Rényi entropy can be written as $H_0(P) = \log|\mathrm{supp}(P)|$, where $\mathrm{supp}(P)$ is the support of $P$: $\mathrm{supp}(P) = \{x \colon P(x) > 0\}$. For $\alpha = 1$, we obtain the Shannon entropy: $H_1(P) = -\sum_{x \in X} P(x) \log P(x)$. For $\alpha = 2$, $H_2(P) = -\log \sum_{x \in \mathcal{X}} P^2(x)$ is the logarithm of the collision probability: $H_2(D) = -\log \Pr_{Y_1, Y_2 \sim P}[Y_1 = Y_2]$. Finally, $H_\infty(P) = -\log \sup_{x \in \mathcal{X}} P(x)$. It can be shown that the Rényi entropy is a non-negative decreasing function of $\alpha$: $H_{\alpha_1}(P) > H_{\alpha_2}(P)$ for $\alpha_1 < \alpha_2$.

Our analysis of the multiple adaptation problem makes use of the *Rényi Divergence* which is parameterized by $\alpha$ as for the Rényi entropy and defined by

$$D_\alpha(P\|Q) = \frac{1}{\alpha - 1} \log \sum_x P(x) \left(\frac{P(x)}{Q(x)}\right)^{\alpha-1}. \quad (2)$$

For $\alpha = 1$, $D_1(P\|Q)$ coincides with the standard relative entropy or KL-divergence. For $\alpha = 2$, $D_2(P\|Q) = \log E_{x \sim P} \frac{P(x)}{Q(x)}$ is the logarithm of the expected probabilities ratio. For $\alpha = \infty$, $D_\infty(P\|Q) = \log \sup_{x \in \mathcal{X}} \frac{P(x)}{Q(x)}$, which bounds the maximum ratio between the two probability distributions. We will denote by $d_\alpha(P\|Q)$ the exponential in base 2 of the Rényi divergence:

$$d_\alpha(P\|Q) = 2^{D_\alpha(P\|Q)} = \left[\sum_x \frac{P^\alpha(x)}{Q^{\alpha-1}(x)}\right]^{\frac{1}{\alpha-1}}. \quad (3)$$

Given a class of distributions $\mathcal{Q}$, we denote by $D_\alpha(P\|\mathcal{Q})$ the infimum $\inf_{Q \in \mathcal{Q}} D_\alpha(P\|Q)$. We will concentrate on the case where $\mathcal{Q}$ is the class of all mixture distributions over a set of $k$ source distributions, i.e., $\mathcal{Q} = \{Q_\lambda : Q_\lambda = \sum_{i=1}^k \lambda_i Q_i, \lambda \in \Delta\}$. It can be shown that the Rényi Divergence is always non-negative and that for any $\alpha > 0$, $D_\alpha(P\|Q) = 0$ iff $P = Q$, (see [1]).

## 3 Multiple Source Adaptation Guarantees

### 3.1 Known Target Distribution

Here, we assume that the target distribution $P$ is known to the learner. We give a general method for determining a multiple source hypothesis with good performance. This consists of computing a mixture $\lambda$ such that $Q_\lambda$ minimizes $D_\alpha(P\|\mathcal{Q})$ and selecting the distribution weighted hypothesis $h_\lambda$ based on the parameter $\lambda$ found. The hypothesis $h_\lambda$ is proven to benefit from the following guarantee:

$$\mathcal{L}_P(h_\lambda, f) \leq (d_\alpha(P\|\mathcal{Q}) \, \epsilon)^{\frac{\alpha-1}{\alpha}} M^{\frac{1}{\alpha}}. \quad (4)$$

Note that in the determination of $\lambda$ we do not use any information regarding the various hypotheses $h_i$. We start with the following useful lemma which relates the average loss based on two different distributions and the Rényi divergence between these distributions.

**Lemma 1** *For any distributions $P$ and $Q$, functions $f$ and $h$ and loss $L$ and $\alpha > 1$, the following inequalities hold:*

$$\mathcal{L}_P(h, f) \leq \left(d_\alpha(P\|Q) E_{x \sim Q}[L^{\frac{\alpha}{\alpha-1}}(h(x), f(x))]\right)^{\frac{\alpha-1}{\alpha}}$$
$$\leq (d_\alpha(P\|Q) \mathcal{L}_Q(h, f))^{\frac{\alpha-1}{\alpha}} M^{\frac{1}{\alpha}}.$$

**Proof:** The lemma follows from the following:

$$\mathcal{L}_P(h, f) = \sum_x \frac{P(x)}{Q^{\frac{\alpha-1}{\alpha}}(x)} Q^{\frac{\alpha-1}{\alpha}}(x) L(h(x), f(x))$$
$$\leq \left[\sum_x \frac{P^\alpha(x)}{Q^{\alpha-1}(x)}\right]^{\frac{1}{\alpha}} \left[\sum_x Q(x) L^{\frac{\alpha}{\alpha-1}}(h(x), f(x))\right]^{\frac{\alpha-1}{\alpha}}$$
$$= (d_\alpha(P\|Q))^{\frac{\alpha-1}{\alpha}} \left[\mathop{E}_{x \sim Q}[L^{\frac{\alpha}{\alpha-1}}(h(x), f(x))]\right]^{\frac{\alpha-1}{\alpha}},$$

where we used Hölder's inequality. The second inequality in the statement of the lemma follows from the upper bound $M$ on the loss $L$. ∎

We now use this result to prove a general guarantee for adaptation with multiple sources.

**Theorem 2** *Consider the multiple source adaptation setting. For any distribution $P$ there is a hypothesis $h_\lambda(x) = \sum_{i=1}^k \frac{\lambda_i Q_i(x)}{Q_\lambda(x)} h_i(x)$, such that*

$$\mathcal{L}_P(h_\lambda, f) \leq (d_\alpha(P\|\mathcal{Q}) \, \epsilon)^{\frac{\alpha-1}{\alpha}} M^{\frac{1}{\alpha}}.$$

**Proof:** Let $Q_\lambda(x) = \sum_{i=1}^k \lambda_i Q_i(x)$ be the mixture distribution that minimizes $D_\alpha(P\|Q_\lambda)$. The average loss of the hypothesis $h_\lambda$ for the distribution $Q_\lambda$ can be bounded as follows,

$$\mathcal{L}_{Q_\lambda}(h_\lambda, f) = \sum_x Q_\lambda(x) L\Big(\sum_i \frac{\lambda_i Q_i(x)}{Q_\lambda(x)} h_i(x), f(x)\Big)$$
$$\leq \sum_x \sum_i \lambda_i Q_i(x) L(h_i(x), f(x))$$
$$= \sum_i \lambda_i \mathcal{L}_{Q_i}(h_i, f) \leq \epsilon,$$

where the first inequality follows from the convexity of $L$. By Lemma 1, this implies that

$$\mathcal{L}_P(h_\lambda, f) \leq (d_\alpha(P\|Q_\lambda) \, \epsilon)^{\frac{\alpha-1}{\alpha}} M^{\frac{1}{\alpha}}. \quad ∎$$

The case where the target distribution is a mixture, i.e., $P \in \mathcal{Q}$, is the special case treated by [12]. Specifically, when $P \in \mathcal{Q}$, then $d_\alpha(P\|\mathcal{Q}) = 1$ for any $\alpha$, in particular, $d_\infty(P\|\mathcal{Q}) = 1$, which implies the following corollary.

**Corollary 3** *Consider the multiple source adaptation setting. For any mixture distribution $P \in \mathcal{Q}$ there exists a hypothesis $h_\lambda(x) = \sum_{i=1}^k \frac{\lambda_i Q_i(x)}{Q_\lambda(x)} h_i(x)$ such that $\mathcal{L}_P(h_\lambda, f) \leq \epsilon$.*

### 3.2 Unknown Target Distribution

This section considers the case where the target distribution is unknown. Clearly, the performance of the hypothesis depends on the target distribution, but here the hypothesis selected is determined without knowledge of the target distribution, and is based only on the source distributions $Q_i$ and the matching hypotheses $h_i$. Remarkably, the generalization bound obtained is very similar to that of Theorem 2. We start with the following theorem of [12].

**Theorem 4 ([12])** *Let $U(x)$ be the uniform distribution over $\mathcal{X}$. Consider the multiple source adaptation setting. For any $\delta > 0$, there exists a function*

$$h_{\lambda,\eta} = \sum_{i=1}^{k} \frac{\lambda_i Q_i(x) + (\eta/k)U(x)}{\sum_{j=1}^{k} \lambda_j Q_j(x) + \eta U(x)} h_i(x),$$

*whose average loss for any mixture distribution $Q_\mu$ is bounded by:* $\mathcal{L}_{Q_\mu}(h_{\lambda,\eta}, f) \leq \epsilon + \delta$.

We shall use this theorem in our setting.

**Theorem 5** *Consider the multiple source adaptation setting. For any $\delta > 0$, there exists a function*

$$h_{\lambda,\eta} = \sum_{i=1}^{k} \frac{\lambda_i Q_i(x) + (\eta/k)U(x)}{\sum_{j=1}^{k} \lambda_j Q_j(x) + \eta U(x)} h_i(x),$$

*whose average loss for any distribution $P$ is bounded by,*

$$\mathcal{L}_P(h_{\lambda,\eta}, f) \leq \big[d_\alpha(P\|Q)(\epsilon + \delta)\big]^{\frac{\alpha-1}{\alpha}} M^{\frac{1}{\alpha}}.$$

**Proof:** Let $Q_\mu$ be the mixture which minimizes $d_\alpha(P\|\mathcal{Q})$. By Lemma 1, the following holds:

$$\mathcal{L}_P(h, f) \leq \big(d_\alpha(P\|Q_\mu)\mathcal{L}_{Q_\mu}(h, f)\big)^{\frac{\alpha-1}{\alpha}} M^{\frac{1}{\alpha}}.$$

Selecting the hypothesis $h_{\lambda,\eta}$ guaranteed by Theorem 4 yields $\mathcal{L}_{Q_\mu}(h_{\lambda,\eta}, f) \leq \epsilon + \delta$. ∎

### 3.3 Lower Bound

This section shows that the bounds derived in Lemma 1, Theorem 2, and Theorem 5 are almost tight. For the lower bound, we assume that all distributions $Q_i$ and hypotheses $h_i$ are identical, i.e., $Q_i = Q$ and $h_i = h$ for all $i \in [1, k]$. This implies that for any $\lambda \in \Delta$ the equalities $Q_\lambda = Q$ and $h_\lambda = h$ (in fact, any "reasonable" combining rule would return $h$). This leads to the following lower bound for a target distribution $P$.

**Theorem 6** *Let $L$ be the 0-1 loss. For any distribution $Q$, Boolean hypothesis $h$, and Boolean target function $f$ such that $\mathcal{L}_Q(h, f) = \epsilon$, for any $\delta_\alpha \geq \frac{1}{\alpha-1}\log(1 + \epsilon)$, there exists a target distribution $P$ such that $D_\alpha(P\|Q) \leq \delta_\alpha$ and $\mathcal{L}_P(h, f) = [2^{(\alpha-1)\delta_\alpha} - 1]^{\frac{1}{\alpha}} \epsilon^{\frac{\alpha-1}{\alpha}}$.*

**Proof:** Given two Boolean functions $h$ and $f$ let $Err$ denote the domain over which they disagree: $Err = \{x: f(x) \neq h(x)\}$. By assumption, $Q(Err) = \epsilon$. Let $r = \big[\frac{2^{(\alpha-1)\delta_\alpha}-1}{\epsilon}\big]^{\frac{1}{\alpha}} \geq 1$. Define the distribution $P$ as follows: for any $x \in Err$, $P(x) = rQ(x)$, and for any $x \notin Err$, $P(x) = \frac{1-r\epsilon}{1-\epsilon}Q(x)$. Observe that $P$ defines indeed a distribution. Furthermore, by construction, $P(Err) = r\epsilon$. We now show that $D_\alpha(P\|Q) \leq \delta_\alpha$.

$$d_\alpha(P\|Q) = \left[\sum_x \frac{P^\alpha(x)}{Q^{\alpha-1}(x)}\right]^{\frac{1}{\alpha-1}}$$

$$= \left[\sum_{x \in Err} \frac{P^\alpha(x)}{Q^{\alpha-1}(x)} + \sum_{x \notin Err} \frac{P^\alpha(x)}{Q^{\alpha-1}(x)}\right]^{\frac{1}{\alpha-1}}$$

$$= \big[r\epsilon(r)^{\alpha-1} + (1-r\epsilon)\big(\tfrac{1-r\epsilon}{1-\epsilon}\big)^{\alpha-1}\big]^{\frac{1}{\alpha-1}}$$

$$\leq (r^\alpha \epsilon + 1)^{\frac{1}{\alpha-1}} = 2^{\delta_\alpha}.$$

which completes the proof. ∎

The lower bound of Theorem 6 is almost tight, when compared to Lemma 1. The ratio between the upper bound (Lemma 1) and the lower bound (Theorem 6) is only $[1 - (d_\alpha(P\|Q))^{-(\alpha-1)}]^{\frac{1}{\alpha}}$. In addition, for $D_\alpha(P\|Q) < \frac{1}{\alpha-1}\log(1 + \epsilon)$, by Lemma 1, we have that $\mathcal{L}_P(h, f) \leq (1+\epsilon)^{\frac{1}{\alpha}}(\epsilon)^{\frac{\alpha-1}{\alpha}}$.

### 3.4 Simple Combining Rules

In this section, we consider a set of "simple" combining rules and derive an upper bound on their loss. These combining rules are simple in the sense that they do not depend at all on the target distribution but only slightly on the source distributions. Specifically, we consider the following family of hypothesis combinations, which we call $r$-*norm combinations*:

$$h_{r\text{-norm}}(x) = \sum_{i=1}^{k} \frac{Q_i^r(x)}{\sum_{j=1}^{k} Q_j^r(x)} h_i(x).$$

The $r$-norm combinations include several natural combination rules. For $r = 1$, we obtain the *uniform combining rule*:

$$h_{uni}(x) = \sum_{i=1}^{k} \frac{Q_i(x)}{\sum_{j=1}^{k} Q_j(x)} h_i(x),$$

which is a distribution weighted combination rule. The value $r = \infty$ gives the *maximum combining rule*,

$$h_{max}(x) = h_{i_{max}}(x) \text{ where } i_{max} = \underset{j}{\operatorname{argmax}}\, Q_j(x).$$

For the $r$-norm combining rules we will make an assumption based on the following definition relating the target distribution $P$ and the source distributions $Q_i$.

**Definition 7** *A distribution $P$ is $(\rho, r)$-norm-bounded by distributions $Q_1, \ldots, Q_k$ if for all $x \in X$ and $r \geq 1$, the following holds:* $P(x) \leq \rho \big[\sum_{i=1}^{k} Q_i^r(x)\big]^{1/r}$.

We can now establish the performance of an $r$-norm hypothesis $h_{r\text{-norm}}$ in the case where $P$ is $(\rho,r)$-norm-bounded by $Q_1,\ldots,Q_k$.

**Theorem 8** *For any distribution $P$ that is $(\rho,r)$-norm-bounded by $Q_1,\ldots,Q_k$, the average loss of $h_{r\text{-norm}}$ is bounded as follows: $\mathcal{L}_P(h_{r\text{-norm}},f) \leq \rho k \epsilon$.*

**Proof:** By the convexity of the loss function $L$,

$$\mathcal{L}_P(h_{r\text{-norm}},f) = \sum_x P(x) L\Big(\sum_i \frac{Q_i^r(x)}{\sum_j Q_j^r(x)} h_i(x), f(x)\Big)$$

$$\leq \sum_x \sum_{i=1}^k Q_i^r(x) \frac{P(x)}{\sum_j Q_j^r(x)} L(h_i(x), f(x))$$

$$= \sum_x \sum_{i=1}^k Q_i(x) \Big(\frac{Q_i^r(x)}{\sum_j Q_j^r(x)}\Big)^{1-\frac{1}{r}} \frac{P(x)}{(\sum_j Q_j^r(x))^{\frac{1}{r}}} L(h_i(x), f(x))$$

$$\leq \sum_x \sum_{i=1}^k Q_i(x) \rho L(h_i(x), f(x)) = \rho \sum_{i=1}^k \epsilon_i \leq \rho k \epsilon,$$

where the second inequality uses the assumption that $P$ is $(\rho,r)$-norm-bounded bounded by $Q_1,\ldots,Q_k$. ∎

The following lemma relates the notion of $(\rho,r)$-norm-boundedness to the Rényi divergence.

**Lemma 9** *For any distribution $P$ that is $(\rho, r-1)$-norm-bounded by $Q_1,\ldots,Q_k$, the following inequality holds:*

$$D_r(P\|Q_u) \leq \log k\rho,$$

*where $Q_u(x) = \sum_{i=1}^k (1/k) Q_i(x)$.*

**Proof:** By definition of $d_r^{r-1}(P\|Q_u)$, we can write

$$d_r^{r-1}(P\|Q_u) = \sum_x P(x) \frac{P^{r-1}(x)}{(\sum_{i=1}^k \frac{1}{k} Q_i(x))^{r-1}}$$

$$= k^{r-1} \sum_x P(x) \frac{P^{r-1}(x)}{(\sum_{i=1}^k Q_i(x))^{r-1}}$$

$$\leq k^{r-1} \sum_x P(x) \frac{P^{r-1}(x)}{\sum_{i=1}^k Q_i^{r-1}(x)}$$

$$\leq k^{r-1} \sum_x P(x) \rho^{r-1} = (k\rho)^{r-1}.$$

Taking the $\log$ gives the bound on the divergence:

$$D_r(P\|Q_u) \leq \frac{1}{r-1} \log(k\rho)^{r-1} = \log k\rho.$$
∎

We can now derive a bound for an arbitrary hypothesis $h$ in the case where $P$ is $(\rho,r)$-norm-bounded by $Q_1,\ldots,Q_k$, as a function of the loss on the individual domains $Q_i$.

**Theorem 10** *For any distribution $P$ that is $(\rho, r-1)$-norm-bounded by $Q_1,\ldots,Q_k$ the following bound holds:*

$$\mathcal{L}_P(h,f) \leq \Big[\rho \sum_{i=1}^k \mathcal{L}_{Q_i}(h,f)\Big]^{\frac{r-1}{r}} M^{\frac{1}{r}}. \quad (5)$$

## 4 Approximate Distributions

This section discusses the case where instead of the true distribution $Q_i$ for source $i$, the learner has access only to an approximation $\widehat{Q}_i$. This is a situation that can arise in practice: a hypothesis $h_i$ is learned by training on a labeled sample drawn from $Q_i$, which is also used to derive a model $\widehat{Q}_i$ for the distribution $Q_i$. As before, we shall assume that the average loss of each hypothesis $h_i$ is at most $\epsilon$ with respect to the original distribution $Q_i$ and deal separately with the cases of a known or unknown target distribution.

### 4.1 Known Target Distribution

We wish to proceed as in Section 3.1, where we determine the parameter $\lambda$ that minimizes the divergence between $P$ and a mixture of the source distributions. However, since here we are only given approximate source distributions, we need to modify that approach as follows: (1) since we only have access to $\widehat{Q}_i$, we shall compute a mixture $\widehat{\lambda} = \operatorname{argmin}_\mu D_\alpha(P\|\widehat{\mathcal{Q}})$ rather than $\lambda = \operatorname{argmin}_\mu D_\alpha(P\|\mathcal{Q})$, where $\widehat{\mathcal{Q}}$ is the set of mixture distributions over $\widehat{Q}_i$; (2) our hypothesis will be based on $\widehat{Q}_i$: $h_\mu(x) = \sum_{i=1}^k \frac{\mu_i \widehat{Q}_i(x)}{\widehat{Q}_\mu(x)} h_i(x)$.

The following lemma relates the divergence of the individual distributions to that of the mixture.

**Lemma 11** *Let $\alpha > 1$. For any $\mu \in \Delta$, the following holds:*

$$D_\alpha(Q_\mu\|\widehat{Q}_\mu) \leq \max_i D_\alpha(Q_i\|\widehat{Q}_i).$$

**Proof:** For $\alpha > 1$ the function $g: (x,y) \mapsto x^\alpha/y^{\alpha-1}$ is convex.[1] Thus, we can write

$$d_\alpha^{\alpha-1}(Q_\mu\|\widehat{Q}_\mu) = \sum_x \frac{Q_\mu^\alpha(x)}{\widehat{Q}_\mu^{\alpha-1}(x)} = \sum_x \frac{[\sum_{i=1}^k \mu_i Q_i(x)]^\alpha}{[\sum_{i=1}^k \mu_i \widehat{Q}_i(x)]^{\alpha-1}}$$

$$\leq \sum_x \sum_i \mu_i \frac{Q_i^\alpha(x)}{\widehat{Q}_i^{\alpha-1}(x)} = \sum_i \mu_i d_\alpha^{\alpha-1}(Q_i\|\widehat{Q}_i)$$

$$\leq \max_i d_\alpha^{\alpha-1}(Q_i\|\widehat{Q}_i).$$
∎

The next lemma establishes a triangle inequality-like property with a slight increase of the parameter $\alpha$.

**Lemma 12** *For any $\alpha > 1$, the following inequality holds:*

$$D_\alpha(P\|\widehat{Q}) \leq D_{2\alpha}(P\|Q) + D_{2\alpha-1}(Q\|\widehat{Q}).$$

**Proof:** By definition of the divergence $D_\alpha$ and by the

---
[1]The convexity of $g$ follows from the semi-definite positiveness of the Hessian. It can be shown that it has one positive and one zero eigenvalue.

Cauchy-Schwartz inequality, the following holds:

$$d_\alpha^{\alpha-1}(P\|\widehat{Q}) = \sum_x \frac{P^\alpha(x)}{\widehat{Q}^{\alpha-1}(x)} = \sum_x \frac{P^\alpha(x)}{Q^{\alpha-1/2}(x)} \frac{Q^{\alpha-1/2}(x)}{\widehat{Q}^{\alpha-1}(x)}$$
$$\leq \sqrt{\sum_x \frac{P^{2\alpha}(x)}{Q^{2\alpha-1}(x)}} \sqrt{\sum_x \frac{Q^{2\alpha-1}(x)}{\widehat{Q}^{2\alpha-2}(x)}}$$
$$= d_{2\alpha}^{\frac{2\alpha-1}{2}}(P\|Q)\, d_{2\alpha-1}^{\frac{2\alpha-2}{2}}(Q\|\widehat{Q}).$$

Taking the log yields

$$(\alpha-1)D_\alpha(P\|\widehat{Q}) \leq (\alpha-\tfrac{1}{2})D_{2\alpha}(P\|Q) + (\alpha-1)D_{2\alpha-1}(Q\|\widehat{Q})$$

and thus

$$D_\alpha(P\|\widehat{Q}) \leq \frac{\alpha-\tfrac{1}{2}}{\alpha-1} D_{2\alpha}(P\|Q) + D_{2\alpha-1}(Q\|\widehat{Q})$$
$$\leq D_{2\alpha}(P\|Q) + D_{2\alpha-1}(Q\|\widehat{Q}),$$

which completes proof of the lemma. ∎

We can now establish the main theorem of this section. The bound presented depends only on the divergence between $P$ and $\mathcal{Q}$ (the mixtures of the true distributions) and the divergence between the approximated distributions $\widehat{Q}_i$ and the true distribution $Q_i$.

**Theorem 13** *Let* $\lambda = \mathrm{argmin}_\mu D_\alpha(P\|Q_\mu)$ *and* $\widehat{\lambda} = \mathrm{argmin}_\mu D_\alpha(P\|\widehat{Q}_\mu)$. *Then,*

$$\mathcal{L}_P(h_{\widehat{\lambda}}, f) \leq \epsilon^{\gamma^2} d_{2\alpha}^\gamma(P\|\mathcal{Q}) M^{\frac{1+\gamma}{\alpha}} \cdot$$
$$\max_i d_{2\alpha-1}^\gamma(Q_i\|\widehat{Q}_i) \max_i d_\alpha^{\gamma^2}(\widehat{Q}_i\|Q_i),$$

*where* $\gamma = \frac{\alpha-1}{\alpha}$.

**Proof:** By Lemma 1, we can write

$$L_P(h_{\widehat{\lambda}}, f) \leq [d_\alpha(P\|\widehat{Q}_{\widehat{\lambda}})]^\gamma \mathcal{L}_{\widehat{Q}_{\widehat{\lambda}}}^\gamma(h_{\widehat{\lambda}}, f) M^{\frac{1}{\alpha}}.$$

By convexity of $L$, $\mathcal{L}_{\widehat{Q}_{\widehat{\lambda}}}(h_{\widehat{\lambda}}, f)$ can be bounded by

$$\sum_{i=1}^k \widehat{\lambda}_i \mathcal{L}_{\widehat{Q}_i}(h_i, f) \leq \sum_{i=1}^k \widehat{\lambda}_i [d_\alpha(\widehat{Q}_i\|Q_i)]^\gamma \mathcal{L}_{Q_i}^\gamma(h_i, f) M^{\frac{1}{\alpha}}$$
$$\leq \epsilon^\gamma M^{\frac{1}{\alpha}} \max_{i=1}^k (d_\alpha(\widehat{Q}_i\|Q_i))^\gamma,$$

where the first inequality uses Lemma 1, and the last one our assumption on the loss of $h_i$. By definition of $\widehat{\lambda}$, the divergence $D_\alpha(P\|\widehat{Q}_{\widehat{\lambda}})$ can be bounded by

$$D_\alpha(P\|\widehat{Q}_\lambda) \leq D_{2\alpha}(P\|Q_\lambda) + D_{2\alpha-1}(Q_\lambda\|\widehat{Q}_\lambda)$$
$$\leq D_{2\alpha}(P\|Q_\lambda) + \max_i D_{2\alpha-1}(Q_i\|\widehat{Q}_i),$$

where the first inequality holds by Lemma 12 and the last one by Lemma 11. The theorem follows from combining the inequalities just derived. ∎

## 4.2 Unknown Target Distribution

In this section we address the case where the target distribution $P$ is unknown, as in Section 3.2. One main conceptual difficulty here is that we are given the distributions $\widehat{Q}_i$, but the assumption on the average loss of the hypothesis $h_i$ holds for $Q_i$, not $\widehat{Q}_i$. Another issue is that we wish to give a generalization bound that depends on the divergence between $P$ and $\mathcal{Q}$, rather than the divergence between $P$ and $\widehat{\mathcal{Q}}$. The following theorem bounds the average loss with respect to an arbitrary mixture of the approximate distributions.

**Theorem 14** *Consider the multiple source adaptation setting where the learner receives access to an approximate distribution $\widehat{Q}_i$ instead of the true distribution $Q_i$ of source $i$. Then, for any $\delta > 0$, there exists an approximate distribution weighted combination hypothesis*

$$h_{\lambda,\eta} = \sum_{i=1}^k \frac{\lambda_i \widehat{Q}_i(x) + (\eta/k)U(x)}{\sum_{j=1}^k \lambda_j \widehat{Q}_j(x) + \eta U(x)} h_i(x),$$

*such that for any mixture distribution* $\widehat{Q}_\mu$,

$$\mathcal{L}_{\widehat{Q}_\mu}(h_{\lambda,\eta}, f) \leq \left[\max_i d_\alpha(\widehat{Q}_i\|Q_i)\,\epsilon\right]^{\frac{\alpha-1}{\alpha}} M^{\frac{1}{\alpha}} + \delta.$$

**Proof:** Let $\widehat{\epsilon}$ denote the maximum average loss $\widehat{\epsilon} = \max_i \mathcal{L}_{\widehat{Q}_i}(h_i, f)$. By Theorem 4, for any $\delta > 0$, there exists a hypothesis $h_{\lambda,\eta}$ such that $\mathcal{L}_{Q_\mu}(h_{\lambda,\eta}, f) \leq \widehat{\epsilon} + \delta$. Now, by Lemma 1, for any $i \in [1, k]$,

$$\mathcal{L}_{\widehat{Q}_i}(h_i, f) \leq \left[d_\alpha(\widehat{Q}_i\|Q_i)\mathcal{L}_{Q_i}(h_i, f)\right]^{\frac{\alpha-1}{\alpha}} M^{\frac{1}{\alpha}}.$$

Since by assumption $\mathcal{L}_{Q_i}(h_i, f) \leq \epsilon$, it follows that

$$\mathcal{L}_{\widehat{Q}_i}(h_i, f) \leq \left[d_\alpha(\widehat{Q}_i\|Q_i)\,\epsilon\right]^{\frac{\alpha-1}{\alpha}} M^{\frac{1}{\alpha}},$$

for all $i \in [1, k]$. Thus, by its definition, $\widehat{\epsilon}$ can be bounded by $[\max_i d_\alpha(\widehat{Q}_i\|Q_i)\,\epsilon]^{\frac{\alpha-1}{\alpha}} M^{\frac{1}{\alpha}}$, which proves the statement of the theorem. ∎

The following corollary is a straightforward consequence of the theorem.

**Corollary 15** *Consider the multiple source adaptation setting where the learner receives access to an approximate distribution $\widehat{Q}_i$ instead of the true distribution $Q_i$ of source $i$. Then, for any $\delta > 0$, there exists an approximate distribution weighted combination hypothesis*

$$h_{\lambda,\eta} = \sum_{i=1}^k \frac{\lambda_i \widehat{Q}_i(x) + (\eta/k)U(x)}{\sum_{j=1}^k \lambda_j \widehat{Q}_j(x) + \eta U(x)} h_i(x),$$

*such that for any distribution $P$,*

$$\mathcal{L}_P(h_{\lambda,\eta}, f) \leq [d_\alpha(P\|\widehat{\mathcal{Q}})(\widehat{\epsilon} + \delta)]^{\frac{\alpha-1}{\alpha}} M^{\frac{1}{\alpha}},$$

*with* $\widehat{\epsilon} \leq [\max_i d_\alpha(\widehat{Q}_i\|Q_i)\epsilon]^{\frac{\alpha-1}{\alpha}} M^{\frac{1}{\alpha}}$, *and*

$$D_\alpha(P\|\widehat{\mathcal{Q}}) \leq D_{2\alpha}(P\|\mathcal{Q}) + \max_i D_{2\alpha-1}(Q_i\|\widehat{Q}_i).$$

## 5 Multiple Target Functions

This section examines the case where the target or labeling functions of the source domains are distinct.

Let $f_i$ denote the target function associated to source $i$. We shall assume that $\mathcal{L}_P(f_i, f) \leq \delta$ for al $i \in [1, k]$, where $f$ is the labeling function associated to the target domain $P$. Note that we require the source functions $f_i$ to be close to the target function $f$ only on the target distribution, and not on the source distribution $Q_i$. Thus, we do not assume that $h_i$ has a small loss with respect to $f$ on $Q_i$.

Here, we shall also assume that the loss function verifies the *triangle inequality*: $L(g_1, g_3) \leq L(g_1, g_2) + L(g_2, g_3)$ for all $g_1, g_2, g_3$, and is convex with respect to both arguments, i.e., $L(\sum_i \mu_i h_i, \sum_i \mu_i f_i) \leq \sum_i \mu_i L(h_i, f_i)$, for all $h_i, f_i$, $i \in [1, k]$, and $\mu \in \Delta$.

**Theorem 16** *Assume that the loss function $L$ is convex and obeys the triangle inequality. Then, for any $\lambda \in \Delta$, the following holds:*

$$\mathcal{L}_P(h_\lambda, f) \leq \left[d_\alpha(P\|Q_\lambda)\epsilon\right]^\gamma M^{\frac{1}{\alpha}} + k\delta,$$

*where $\gamma = \frac{\alpha-1}{\alpha}$.*

**Proof:** Let $f_\lambda(x) = \sum_{i=1}^k \lambda_i Q_i(x) f_i(x)/Q_\lambda(x)$. Observe that by convexity of $L$,

$$\mathcal{L}_P(f_\lambda, f) \leq \sum_{i=1}^k \sum_x \frac{\lambda_i Q_i(x)}{Q_\lambda(x)} P(x)[L(f_i(x), f(x))]$$
$$\leq \sum_{i=1}^k \sum_x P(x)[L(f_i(x), f(x))] \leq k\delta.$$

Thus, by the triangle inequality, and Lemma 1,

$$\mathcal{L}_P(h_\lambda, f) \leq \mathcal{L}_P(h_\lambda, f_\lambda) + \mathcal{L}_P(f_\lambda, f)$$
$$\leq (d_\alpha(P\|Q_\lambda)\mathcal{L}_{Q_\lambda}(h_\lambda, f_\lambda))^\gamma M^{\frac{1}{\alpha}} + k\delta$$
$$\leq \left(d_\alpha(P\|Q_\lambda) \sum_{i=1}^k \lambda_i \mathcal{L}_{Q_i}(h_i, f_i)\right)^\gamma M^{\frac{1}{\alpha}} + k\delta$$
$$\leq (d_\alpha(P\|Q_\lambda)\epsilon)^\gamma + k\delta,$$

where the third inequality follows from the convexity of $L$ and the last inequality holds by the bound assumed on the expected loss of each source hypothesis $h_i$. ∎

A similar bound can be given in the case where the loss verifies only a relaxed version of the triangle inequality ($\beta$-inequality): $L(g_1, g_3) \leq \beta(L(g_1, g_2) + L(g_2, g_3))$, for all $g_1, g_2, g_3$ for some $\beta > 0$.

**Theorem 17** *Assuming that the loss $L$ is convex and verifies the $\beta$-inequality, then for any $\lambda \in \Delta$, the following bound holds:*

$$\mathcal{L}_P(h_\lambda, f) \leq \beta\left[d_\alpha(P\|Q_\lambda)\epsilon\right]^{\frac{\alpha-1}{\alpha}} M^{\frac{1}{\alpha}} + \beta k\delta.$$

## 6 Experiments

This section presents an empirical evaluation of the distribution weighted combination rule based on both artificial and real-world data.

**Artificial Data:** Here, we created a two-dimensional artificial dataset using four Gaussians distributions $[g_1, g_2, g_3, g_4]$ with means $[(1,1), (-1,1), (-1,-1), (1,-1)]$ and unit variance. The source distributions $Q_1$ and $Q_2$ were generated from the uniform mixture of $[g_1, g_2, g_3]$ and $[g_1, g_3, g_4]$, respectively, and the target distribution $P$ was generated from the uniform mixture of $[g_1, \ldots, g_4]$. The labeling function was defined as $f(x_1, x_2) = \text{sign}(x_1 x_2)$. For training and testing, we sampled 5,000 points from each distribution. Note that $P = \frac{1}{4}(g_1 + \ldots + g_4)$ cannot be constructed with any mixture $\lambda Q_1 + (1-\lambda)Q_2 = \frac{1}{3}(g_1 + \lambda g_2 + g_3 + (1-\lambda)g_4)$. Also, note that the base hypotheses, when tested on $P$, misclassify all the points that fall into at least one quadrant of the plane. However, with the use of a distribution weighted combination rule, the appropriate base hypothesis is selected depending on which quadrant a point falls into, and this pitfall is avoided.

We used libsvm (http://www.csie.ntu.edu.tw/~cjlin/libsvm/) with linear kernels to produce base classifiers. We report the mean squared error (MSE) of the resulting (non-thresholded) combination rules. The mean and standard deviation reported are measured over 100 randomly generated datasets. Figure 1(a) shows that the curve plotting the error as a function of the mixture parameter $\lambda$ has the same shape as the Réyni divergence curve, as predicted by our bounds. Note that for $\lambda = 0$ and $\lambda = 1$ we obtain the two basic hypotheses.

**Real-World Data:** For the real-world experiments, we used the sentiment analysis dataset of [4] also used in [12], which consists of product review text and rating labels taken from four different categories: *books* ($B$), *dvds* ($D$), *electronics* ($E$) and *kitchen-wares* ($K$). Using the methodology of [12], we defined a vocabulary of 3,900 words that fall into the intersection of all four domains and occur at least twice. These words were then used to train a bigram statistical language model for each domain using the GRM library (http://www.research.att.com/fsmtools/grm). The same vocabulary was then used to encode each data point as a 3,900-dimensional vector containing the number of occurrences of each word.

In the same vein as the artificial setting, we defined $Q_1$ and $Q_2$ as the uniform mixture of $[E, K, D]$ and $[E, K, B]$, respectively, and the target distribution $P$ as the uniform mixture of $[E, K, D, B]$. Each base hypothesis was trained with 2,000 points using support vector regression (SVR) [18], also implemented by libsvm, and the mixture was evaluated on a test set of 2,666 points. The experiment was repeated 100 times with random test/train splits. Although each base domain in this setting is relatively powerful, we still see a significant improvement when using the distribution weighted combination, as shown in Figure 1(b).

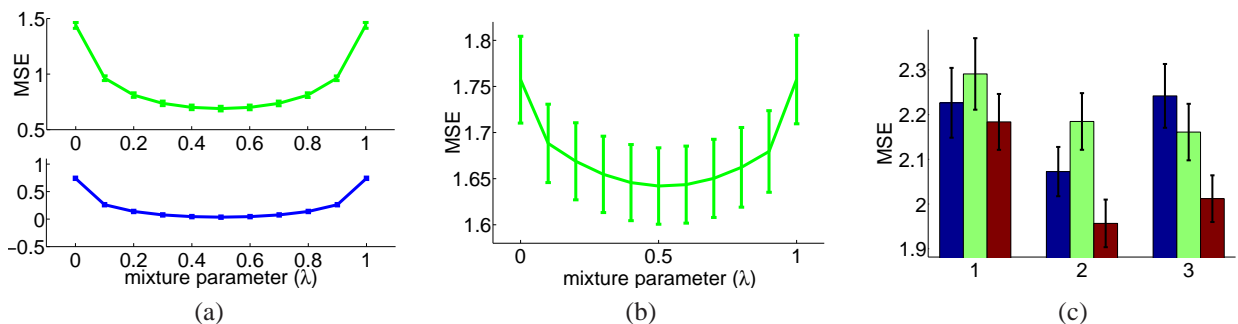

Figure 1: (a) Performance of the distribution weighted combination rule for an artificial dataset, plotted as a function of the mixture parameter $\lambda$; comparison with the Réyni divergence plotted for the same parameter. (b) MSE of the distribution weighted combination rule for the sentiment analysis dataset. (c) MSE of base hypotheses and distribution weighted combination. For each group, the first two bars indicate the MSE of the base hypotheses followed by that of the distribution weighted hypothesis. The base domains were $D$ and $B$ with target domain mixture $K/E$ for group 1; $E$ and $B$ with target $K/D$ for group 2; and $D$ and $E$ with target $B/K$ for group 3.

In a final set of experiments, we trained each of two base hypotheses with $1,000$ points from a single domain. We then tested on a target that is a uniform mixture of the two other domains, consisting of $2,000$ points. Clearly, the target is not a mixture of the base domains. These experiments were repeated 100 times with random test/train splits. As shown in Figure 1(c), and as the caption explains in detail, the distribution weighted combination is capable of doing significantly better than either base hypothesis.

## 7 Conclusion

We presented a general analysis of the problem of multiple source adaptation. Our theoretical and empirical results indicate that distribution weighted combination methods can form effective solutions for this problem, including for real-world applications. Our analysis of approximated distribution case and multiple labeling functions cases help cover other related adaptation problems arising in practice. The family of Rényi divergences naturally emerges in our analysis as the "right" distance between distributions in this context.

## References

[1] C. Arndt. *Information Measures: Information and its Description in Science and Engineering*. Signals and Communication Technology. Springer Verlag, 2004.

[2] S. Ben-David, J. Blitzer, K. Crammer, and F. Pereira. Analysis of representations for domain adaptation. In *Proceedings of NIPS 2006*. MIT Press, 2007.

[3] J. Blitzer, K. Crammer, A. Kulesza, F. Pereira, and J. Wortman. Learning bounds for domain adaptation. In *Proceedings of NIPS 2007*. MIT Press, 2008.

[4] J. Blitzer, M. Dredze, and F. Pereira. Biographies, Bollywood, Boom-boxes and Blenders: Domain Adaptation for Sentiment Classification. In *ACL*, 2007.

[5] K. Crammer, M. Kearns, and J. Wortman. Learning from multiple sources. *JMLR*, 9:1757–1774, 2008.

[6] H. Daumé III and D. Marcu. Domain adaptation for statistical classifiers. *Journal of Artificial Intelligence Research*, 26:101–126, 2006.

[7] M. Dredze, J. Blitzer, P. P. Talukdar, K. Ganchev, J. Graca, and F. Pereira. Frustratingly Hard Domain Adaptation for Parsing. In *CoNLL*, 2007.

[8] J.-L. Gauvain and Chin-Hui. Maximum a posteriori estimation for multivariate gaussian mixture observations of Markov chains. *IEEE Transactions on Speech and Audio Processing*, 2(2):291–298, 1994.

[9] F. Jelinek. *Statistical Methods for Speech Recognition*. The MIT Press, 1998.

[10] J. Jiang and C. Zhai. Instance Weighting for Domain Adaptation in NLP. In *Proceedings of ACL 2007*, pages 264–271, Prague, Czech Republic, 2007.

[11] C. J. Legetter and P. C. Woodland. Maximum likelihood linear regression for speaker adaptation of continuous density hidden markov models. *Computer Speech and Language*, pages 171–185, 1995.

[12] Y. Mansour, M. Mohri, and A. Rostamizadeh. Domain adaptation with multiple sources. In *NIPS 2008*, 2009.

[13] A. M. Martínez. Recognizing imprecisely localized, partially occluded, and expression variant faces from a single sample per class. *IEEE Trans. Pattern Anal. Mach. Intell.*, 24(6):748–763, 2002.

[14] S. D. Pietra, V. D. Pietra, R. L. Mercer, and S. Roukos. Adaptive language modeling using minimum discriminant estimation. In *HLT '91: Proceedings of the workshop on Speech and Natural Language*, pages 103–106, 1992.

[15] B. Roark and M. Bacchiani. Supervised and unsupervised PCFG adaptation to novel domains. In *HLT-NAACL*, 2003.

[16] R. Rosenfeld. A Maximum Entropy Approach to Adaptive Statistical Language Modeling. *Computer Speech and Language*, 10:187–228, 1996.

[17] L. G. Valiant. *A theory of the learnable*. ACM Press New York, NY, USA, 1984.

[18] V. N. Vapnik. *Statistical Learning Theory*. Wiley-Interscience, New York, 1998.